# A Computational Study on Emotions and Temperament in Multi-Agent Systems

Daria Barteneva, Nuno Lau and Luís Paulo Reis[1]

**Abstract.** Recent advances in neurosciences and psychology have provided evidence that affective phenomena pervade intelligence at many levels, being inseparable from the cognition-action loop. Perception, attention, memory, learning, decision-making, adaptation, communication and social interaction are some of the aspects influenced by them. This work draws its inspirations from neurobiology, psychophysics and sociology to approach the problem of building autonomous robots capable of interacting with each other and building strategies based on temperamental decision mechanism. Modelling emotions is a relatively recent focus in artificial intelligence and cognitive modelling. Such models can ideally inform our understanding of human behavior. We may see the development of computational models of emotion as a core research focus that will facilitate advances in the large array of computational systems that model, interpret or influence human behavior. We propose a model based on a scalable, flexible and modular approach to emotion which allows runtime evaluation between emotional quality and performance. The results achieved showed that the strategies based on temperamental decision mechanism strongly influence the system performance and there are evident dependency between emotional state of the agents and their temperamental type, as well as the dependency between the team performance and the temperamental configuration of the team members, and this enable us to conclude that the modular approach to emotional programming based on temperamental theory is the good choice to develop computational mind models for emotional behavioral Multi-Agent systems.

## 1 INTRODUCTION

Emotions are part of our every day lifes. They help us focus attention, remember, prioritize, understand and communicate. The possibility of computation of emotions has interested researchers for many years. The emotions influence decision-making processes, socialization, communication, learning and many other important issues of our life. Implementation of emotions in an artificial organism is an important step for different areas of intervention, since academical inquiry [1-10], education [13-15], communication [11, 16], entertainment and others [12, 17-19, 29, 30]. Researchers have focused on the functions of emotion for computational models trying to describe some of behavioral responses to reinforcing signals, communications which transmit the internal states or social bonding between individuals, which could increase fitness in the context of evolution. Among some models of emotions that are described through the computational process exists different approaches to the proper concept of emotion. Each model results of the definition that is given to the emotional process. Since analysis of needs/satisfactions of the human being [24, 25], passing through the analysis of characteristics of the superior nervous system [26, 28], physiological changes [23, 31], neurobiological processes [27], appraisal mechanism and analysis of the psychology of individual personality [20, 21].

The most important questions we made in this project are: what is emotion? How can we represent emotions through computational model? Many authors have tried to categorized emotions. As Marvin Minsky said [22] our culture sees emotions as a deep and ancient mystery. He also reinforce that the psychology has not even reached a consensus on which emotions exists. He describes emotions as a complex rule schemes which we develop during our mental grow submitted to external influences like learning. A more detailed approach was made by António Damasio [27] who studied emotions from a neurobiological perspective. He divided emotions in two groups: primary and secondary. In first group he included emotions which depends on our limbic system and these are innate reactions. In the second group Damásio included emotions we develop during our life based on our experience. Finally he defined emotions as a process of mental evaluation, simple or complex, with disposal responses directed to the body and the brain resulting in additional brain alterations.

For our project we define emotions as a set of external and internal responses which depends on the set of rules based on agent beliefs, desires and intentions. To proceed with development of our emotional model we need to use some kind of quantitative measure to evaluate emotional state. But what is this "Emotional state"? Interesting definition was given by Mehrabian [21] for this concept. He defined it as transitory conditions of the organism – conditions that can vary substantially, and even rapidly, over the course of a day. He also defined "emotional traits" (i.e. Temperament) as conditions that are stable over periods of the year or even a lifetime. As described in Pavlov's theory [28], all human and animal behaviors are coordinated by the Central Nervous System (CNS). Therefore we can't study emotional agents without considering the particularities of the CNS and, consequently, the particularities of temperamental theory.

[1] A manuscript submission in January 8, 2007.
Luís Paulo Reis is Auxiliar Professor at the Faculty of Engineering of the University of Porto and Researcher at LIACC (NIAD&R) - Artificial Intelligence and Computer Science Lab. (Distributed Artificial Intelligence and Robotics Unit), Portugal (phone: +351 22 5081829; fax: +351 22 5081443; e-mail: lpreis@fe.up.pt).
Nuno Lau is Auxiliar Professor at the Department of Electronics and Telecommunications of the Aveiro University and Researcher at IEETA – Institute of Electronic Engineering and Telematics of Aveiro, Portugal (phone: +351 34 370500; fax: +351 34 370545; e-mail: lau@det.ua.pt).
Daria Barteneva is Master of Science Student at the Faculty of Engineering of the University of Porto, Portugal (e-mail: daria.barteneva@gmail.com).

The classical definition for "Temperament" follows: it is a specific feature of Man, which determines the dynamics of his mental activity and behaviour. Two basic indexes of the dynamics of mental processes and behaviours at present are distinguishable: activity and emotionality. In this project we will analyze and develop an emotional model for the agents with temperament. We will use a complex approach to emotion/temperament concepts: based on physiological (CNS) characteristics and on psychological characteristics of the agents.

Our computational model of emotion is inspired on appraisal theory and on superior nervous system characteristics. Most appraisal theories [32, 33] assume that beliefs, desires and intentions are the basis of reasoning and thus of emotional evaluation of the agents situation. In order to create a more flexible and efficient emotion-based behavior system, the appraisal model is implemented in mixture with Pavlov's temperamental theory [28] which studies the basic reasons for different temperamental behaviors and Eysenck's [26] neurophysiological interpretation of the basic measurements of temperament.

We choose to mix these different theories to study the agent teams and to evaluate the team performance when we define different temperamental configuration of the team.

A simulation environment based on a Cyber-Mouse [34, 35] competition simulator was used to test and evaluate the strategies used on the work. A set of robotic experiments was conducted in order to test the performance of the system.

The paper is organized as follows. Section 2 presents the Emotions & Temperament approach distinguishing the physiological and psychical perspectives to the subject of our study in order to explain the fundamentals for the implementation performed for this project. Here we present Pavlov's theory about superior nervous system, Eysenck's scale and Mehrabian's PAD model. Section 3 describes the simulated environment (Cyber-Mouse) we used. Section 4 describes implementation of physiological and psychical layers we have defined for this project. Section 5 describes describes the evaluation experiences we performed to validate our work. Finally Section 6 presents the conclusion we made and future work.

## 2 EMOTIONS AND TEMPERAMENT

As we already have refered, for constructing our emotional model we studied two subjects: emotional states which characterize the immediate emotional condition of the agent and emotional trait (temperament) which define the personality characteristics and behaviors of the agent and influence his emotional state changes. We decided to approach the study of emotions from different perspectives: physiological and psychical, creating double layer architecture for emotional model to increase the system performance. Let us examine each perspective of our approach.

## A General framework for describing central nervous system and Pavlov's theory

In order to find physiological reasons to emotions appearance we started by studying the Pavlov's theory [28] about superior nervous system activity. Activity can be expressed in different degrees of tendency to act, to participate in the diverse challenges. It is possible to note two extremes: from one side, high energy, fervency and swiftness in the mental activity, the motions and the speech, while another - passiveness, sluggishness, the apathy of mental activity, motion and speech. The second index is dynamicity and it is expressed in different degrees of emotional excitability, in the velocity of appearance and the force of the emotions of man, and in the emotional sensitiveness (receptivity to the emotional actions). Based on this characteristics four basic forms of temperament may be distinguished, which were named as follows: sanguine (living), phlegmatic (slow, calm), choleric (energetic, passionate) and melancholic (locked, inclined to the deep experiences).

The definite scientific explanation of temperaments was given by Ivan Pavlov's theory about the types of higher nervous activity. Pavlov described three properties of the processes of excitation and braking [37]:

- The **force** of the processes of excitation and braking;
- The **steadiness** of the processes of excitation and braking;
- The **mobility** of the processes of excitation and braking.

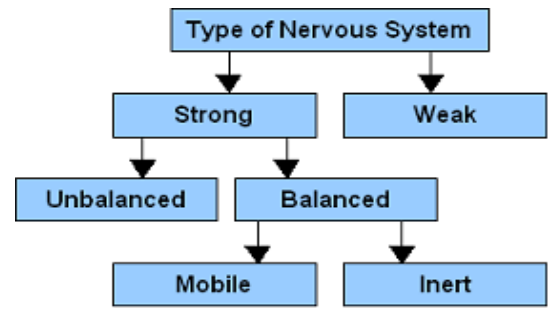

**Figure 1.** Classification of higher nervous system

The combinations of the properties of nervous processes indicated were assumed as basis to determinations of the type of higher nervous activity. Depending on the combination of force, mobility and steadiness of the processes of excitation and braking four basic types of higher nervous activity are distinguished.

Pavlov correlated the types of nervous systems with the psychological types of temperaments isolated with it and revealed their complete similarity. Thus, temperament is a manifestation of the type of nervous system into the activity. As a result the relationship of the types of nervous system and temperaments appears as follows (fig. 1):

- Strong, balanced, mobile type - sanguine temperament;
- Strong, balanced, inert type - phlegmatic

temperament;
- Strong, unbalanced, with the predominance of excitation - choleric temperament;
- Weak type - melancholic temperament.

## Eysenck methodology

One of the things Pavlov tried with his dogs [37] was conflicting conditioning - ringing a bell that signalled food at the same time as another bell that signalled the end of the meal. Some dogs took it well, and maintain their cheerfulness. Some got angry and barked like crazy. Some just laid down and fell asleep. And some whimpered and whined and seemed to have a nervous breakdown.

Pavlov believed that he could account for these personality types with two dimensions: On the one hand there is the overall level of arousal (called excitation) that the dogs' brains had available. On the other, there was the ability the dogs' brains had of changing their level of arousal - i.e. the level of inhibition that their brains had available.
- Lots of arousal, but good inhibition: sanguine.
- Lots of arousal, but poor inhibition: choleric.
- Not much arousal, plus good inhibition: phlegmatic.
- Not much arousal, plus poor inhibition: melancholy.

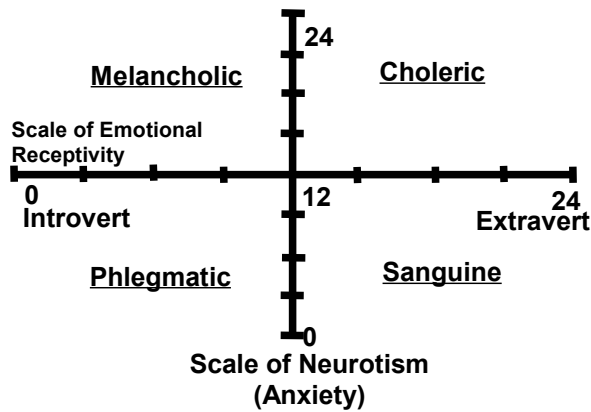

**Figure 2.** Scale of temperamental classification by Eysenck

Arousal would be analogous to warmth, inhibition analogous to moisture. This became the inspiration for Hans Eysenck's theory [26].

In the works of Eysenck a neurophysiological interpretation of the basic measurements of temperament was given, among which were separated - the factor of extraversion -introversion and the factor of neuroticism.

Using the methodology of Eysenck [26] we can perform the personality test to describe the temperament of the individuals by Introvert/Extravert characteristic and Anxiety (Fig. 2).

## Analysis of personality factors in terms of the PAD temperamental model

Analysis of emotional states leads to the conclusion that the human emotions such as anger, fear, depression, elation, etc. are discrete and we need to define some kind of measures to have a basic framework to describe each emotional state using the same scale. After studing the appraisal theory we find Mehrabian model [20, 21] more suitable for computational needs since it defines three dimensions to describe each emotional state and provides an extensive list of emotional labels for points in the PAD space (Fig 3) gives an impression of the emotional meaning of combinations of Pleasure, Arousal and Dominance (PAD).

The three dimensions of the PAD temperament model define a three-dimensional space where individuals are represented as points, personality types are represented as regions and personality scales are represented as straight lines passing through the intersection point of the three axes. Mehrabian uses +P, +A and +D to refer pleasant, arousable and dominant temperament. Respectively, and by using -P,

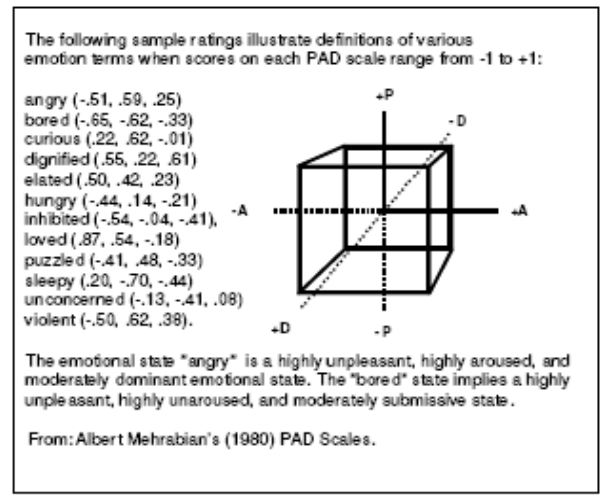

**Figure 3.** Mehrabian PAD temperamental scale.

-A and -D to refer unpleasant, unarousable and submissive temperament, respectively. Since most personality scales load on two or more of the PAD temperament dimensions, Mehrabian define them using the four diagonals in PAD space as follow:
- Exuberant (+P+A+D) vs Bored (-P-A-D)
- Dependent (+P+A-D) vs Disdainful (-P-A+D)
- Relaxed (+P-A+D) vs Anxious (-P+A-D)
- Docile (+P-A-D) vs Hostile (-P+A+D)

In the Analysis of Big-Five Personality factors in terms of PAD temperamental model [38] Mehrabian find the relationship between five temperamental types and the PAD scale. He describe this relationship using linear regressions. The resulting equations are given below for standardized variables with a 0.05 significant level:

Extraversion =        0.24P              +0.72D
Agreeableness =       0.76P   +0.17A    -0.19D
Conscientiousness =   0.29P              +0.28D
Emotional Stability = 0.50P   -0.55A
Sophistication =                +0.28A   +0.60D

*Big Five personality factors*

The Big Five factors and their constituent traits can be summarized as follows:
- *Openness to Experience (or Sophistication)* - Appreciation for art, emotion, adventure, unusual

ideas; imagination and curiosity.
- *Conscientiousness* - A tendency to show self-discipline, act dutifully, and aim for achievement (spontaneousness vs planned behavior).
- *Extraversion* - Energy, surgency, and the tendency to seek stimulation and the company of others.
- *Agreeableness* - A tendency to be compassionate and cooperative rather than suspicious and antagonistic towards others (individualism vs cooperative solutions).
- *Neuroticism* (or *Emotional Stability*) - A tendency to easily experience unpleasant emotions such as anger, anxiety, depression, or vulnerability (emotional stability to stimuli).

We will use this result to determine the emotional state of the agents depending on their temperamental type.

## 3 CYBER-MOUSE SIMULATION ENVIRONMENT

We choose the Cyber-Mouse Simulation environment to implement and test our model because it offers an open, modular and flexible platform permitting unlimited applications and a fully configurable simulation system.

Cyber-Mouse is a modality included in the Micro-Mouse competition organized by Aveiro University (Portugal). This modality is directed to teams interested in the algorithmic issues and software control of mobile autonomous robots. This modality is supported by a software environment, which simulates both robots and a labyrinth [34].

The simulation system possesses a distributed architecture where some types of applications communicate among each other, nominated, a simulator, an application for each agent and a viewer application. The architecture is client-server, where the simulator acts as the server and both the agents and the viewer, acts as clients. This architecture is similar to the Simulation League of RoboCup [36].

The simulator shapes all the components of the robots hardware and the labyrinth. The simulation is executed in discrete time, cycle by cycle. In the beginning of each cycle of simulation the simulator sends to all robotic agents in test, the measures of its sensors, and to all viewers the positions and robots information. The agents can answer with the power values to apply to the engines that command the wheels.

All robots in test have the same physiological characteristics. All have the same sensors and the same engines.

Each robot (fig. 4) is equipped with the following sensors [34, 35]:
- 3 sensors of proximity guided to the front and 60° for each side.
- Beacon sensor that indicates which is the difference between robots direction and the beacon direction.
- Ground sensor, active when robot enters in the arrival zone.
- Compass sensor that allows robot to know which its absolute orientation in the labyrinth is.
- Collision sensor, asset in the case of robot collision.
- Vision sensor, works by identifying other robots and their emotional state.

The measures of the sensors include some noise added for the shape simulator in order to simulate real sensors.

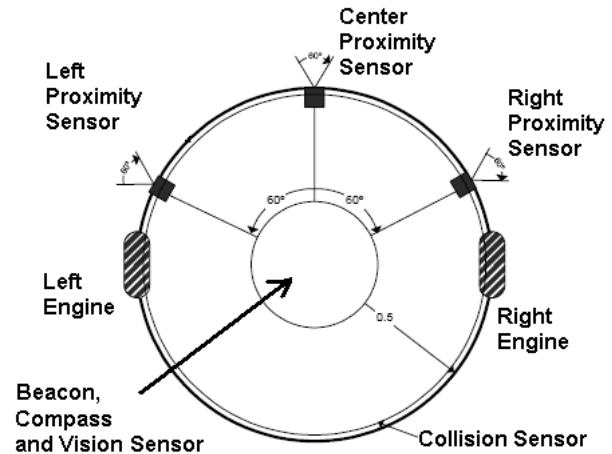

**Figure 4.** Virtual Agent Diagram

In order to detect the beginning of the test and possible interruptions each robot has 2 buttons:
- Start, active when is initiated the test.
- Stop, active when a test interruption exists.

In terms of virtual engines robots is constituted by:
- 2 wheels for 2 independent motors, one on the left and one on the right;
- LED of finishing, to light when reached the arrival zone.

In each cycle of simulation the agents receive the values measured by all its sensors and must decide which power to apply in each motor. The perception that a robotic agent has from the exterior environment is limited and noisy transforming him into the most appropriate tool to perform our work with almost realistic precision.

## 4 A DUAL LAYER MODEL OF EMOTION

As we reference in previous chapters, we choose the approach to emotional programming through the implementation of artificial personalities and the integration of the emotional decision model based on the appraisal theory. The innovation of our approach consists in the duality of our emotional character: it processes the information and gives the output using two different engines, physiological and psychological. In our model the temperament of the agent is defined as the configuration of his mechanical engines and the personality functions which simulates his psyche as the decision mechanism. The emotional response of the agent possesses a dual mechanism: it is physiological (such as motor and sensor force, face expression, mobility) and psychical (such as a vector which defines his internal emotional state).

We also need to emphasize the difference between the agent's temperamental state and agent's emotional state. Temperament, as we already defined, is the steady characteristics of the agent which is "innate" and do not suffer alterations during the agent's life. On the other side, the emotional state of the agent is the dynamic set of values which depends on the external influences, and on the agent's temperament.

We can define emotion as a short episode triggered by an (internal/external) event composed of
- subjective feelings
- inclinations to act
- facial expressions
- cognitive evaluation and
- physiological arousal.

And emotions have a role of heuristic relating events to goals, needs, desires, beliefs of an agent and evaluate their personal relevance and help decision-making.

So, for instance, two agents with different temperaments and the same emotional states on some temporal period, which receive the same external input will have different responses on both, the physiological and the psychical mechanism. We also define different sets of needs and motivations for each temperamental type by the influence of the agent's performance and stimuli on the team work. This modular, but complementary approach, is the core of the innovation of our emotional system and our aspiration of its usability.

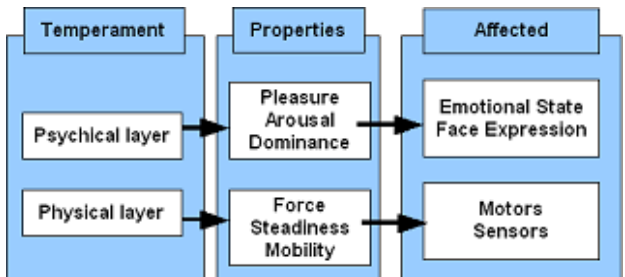

**Figure 5.** Temperamental architecture.

We assume a two layer architecture (Fig. 5) for our emotional model. One layer is physiological and describes the superior Nervous system from the Pavlov perspective. The other layer is psychical and works with the appraisal model created by Mehrabian.

In our temperamental architecture we have not implemented any of dependency between physiological and psychical layers and we are trying to discover some kind of influence that one layer could have on the other layer through the temperamental configurations or common goals implementation. Psychical layer controls the emotional state of the agent through PAD values, and the physiological layer control the engine configuration (motors, sensors, etc...) and the group interaction, based on temperamental needs of the agent (like extroversion/introversion or emotional stability).

Fig. 6 presents the diagram which describes the relationship between the Simulation environment, Decision layer and Data layer. Decision layer works in order to process the inputs received by the agents from simulation environment, determine the outputs that agent return and update his emotional state. Physiological Bank contains the fuzzy measures of force, mobility and steadiness. Values for motors and sensors are archived in Physiological Bank for each temperamental type.

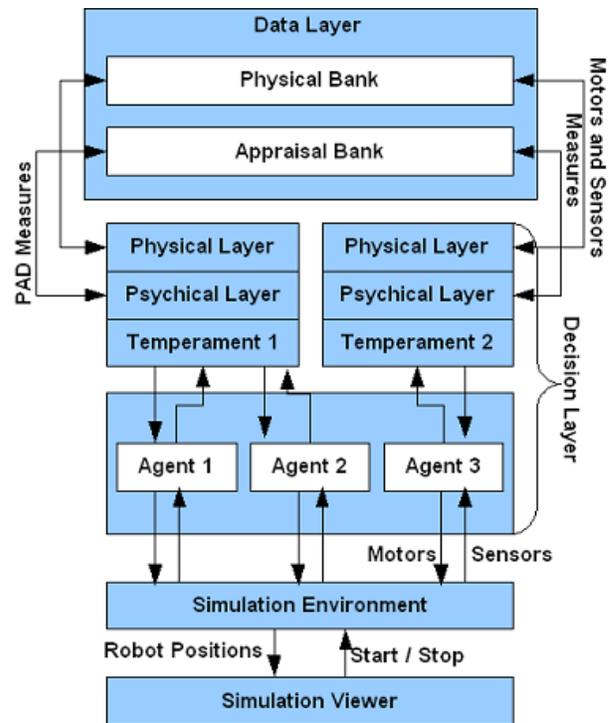

**Figure 6.** Cyber-Mouse Simulation Architecture vs Appraisal Model and Central Nervous System Mechanism

This figure presents a scheme for two temperaments and three agents, but as we already explain we use four temperaments for physiological layer and 5 temperaments for psychical layer:

- Choleric, Sanguine, Phlegmatic and Melancholic defined in **physiological layer** with different Fuzzy set's of values for Force, Steadiness and Mobility.
- Extraversion, Agreeableness, Conscientiousness, Emotional Stability and Sophistication in **psychical layer** which are described using Pleasure, Arousal and Dominance. In order to simplify our model we are using just Extraversion and Emotional Stability in this project.

## Physiological layer

As we show on previous chapter, Pavlov's theory defines the temperamental model based on characteristics of the superior nervous system, but at the same time there are no pure temperamental types in nature, but there are mixtures of different properties which characterize one or another unique temperamental type. So, as we see, one person can have all temperamental types in different ratios. The different proportion of values: force, mobility and steadiness of processes of excitation and braking defines the unique temperamental type for each person. Based on this uncertainty we use Fuzzy Logic to describe and monitorize the temperamental types in our project [39]. In the beginning of the simulation we generate the values which will define the unique combination of temperamental type of the agent, but then these characteristics are changing in run-time in order to adapt the agent state to the external influences. We define the fuzzy intervals for each temperamental variable which define the temperamental characteristics (Force, Mobility, ...) and the value of this variable increases in stressful situations (close threat, wall-shock, etc...) and decreases in calm situations. The speed of the increase and decrease depends on agent's Arousal.

*Force*

In our multi-agent system the force of excitation and braking processes is represented by the force of the motor and reach of the sensors. We define the superior limit for the force value in order to obtain a better simulation of the real world.

*Mobility*

The mobility of the agent is represented by its "persistence" to reach the goal and avoid negative emotions. For instance if some agent is "comfortable" in some place, and his mobility is low, he will not look to move to search other places. He will slow his motors and just stay in the same place until the environment changes forces him to move quickly. At the same time, one agent who has a high mobility will search new places and new directories even if he is comfortable enough in some temporal phase. According to Mehrabian [21] arousal is highly correlated with activity and alertness so changing the Arousal we can control the Mobility of the agent.

*Steadiness*

The steadiness of the agent is the velocity of his emotional state variation. For example, more balanced agents have a slow variation of emotional state. For this we introduce the variable called Anxiety which is used to increase or decrease the Pleasure variable. The value of Anxiety depends on the temperament of the agent. We choose the values for anxiety based on the Eysenck test [26].

*Emotional receptivity*

This variables were based on the Eysenck test described on the second section. The Melancholic and Phlegmatic temperamental types are included in the Introverts group and the Sanguine and Choleric types are included in the Extroverts group. We will evaluate how they performance to reach the beacon, conditioned by their temperamental needs.

## Psychical layer

Our approach does not prescribe a specific set of appraisal dimensions. We have chosen the Pleasure, Arousal, Dominance (PAD) personality-trait and emotional-state scales by Albert Mehrabian [21] because these dimensions are generally not considered to be appraisal dimensions. He argues that any emotion can be expressed in terms of values on these three dimensions, and provides extensive evidence for this claim [20]. This makes his three dimensions suitable for a computational approach. Mehrabian also provides an extensive list of emotional labels for points in the PAD space [21] and gives an impression of the emotional meaning of the combinations of Pleasure, Arousal and Dominance. The emotional-state of an agent can thus be understood as a continuously moving point in an n-dimensional space of appraisal dimensions.

*Appraisal Banks*

The appraisal bank defines the needs, motivations and stimulus of the agent as a set of subjective measures, called appraisal dimensions. First, a simple instrumentation based on appraisal bank that emotionally evaluates events related to survival. Second, a more complex instrumentation based on two appraisal banks, one related to survival the other related to reach the beacon and satisfy temperamental needs. In both banks we have used event-encoding to simulate emotional meaning of events. We now describe how events are interpreted by the two appraisal banks.

As we work with BDI agent's, their thinking are based on **beliefs**, **desires** and **intentions**. Their possess basics to which appraisal based emotions can be added. Lets describe beliefs, desires and intentions of the agents in our simulation system:

Beliefs:

- angry agents are dangerous;
- wall collisions are painful;
- happy agents are friendly and nice;

Desires:

- reach the beacon;
- satisfies personal (temperamental) need like necessity of company of other agents or necessity of loneliness;
- don't get hurt;

Intentions:

- avoid threats (angry agents);
- avoid wall collisions;
- follow happy agents;

We define **Pleasure** as the *conductance of the goal*. For instance if the agent sees the beacon and no obstacle is present his pleasure is high, while if he sees a threat or looses the goal this is highly unpleasant. **Arousal** is the *amount of attention each event needs*, for instance to avoid threats the attention of the agent is needed and lose it needs no attention. **Dominance** is a measure that defines the *amount of freedom of the agent*. For example, if it sees a wall the dominance decreases but if it sees no obstructed way to the goal the dominance increases. It is not a subject of this paper the detailed definition of the appraisal banks.

Appraisal-results are integrated using following formula,

$$E_{t+1} = E_t + \sum_{i=0}^{n} \Delta PAD_{ti}$$

where $E_t$ is the emotional-state at time t, $E_{t+1}$ is the new emotional-state, n is the number of appraisal banks and $\Delta PAD_{ti}$ the appraisal-result vector of bank *i* at time *t*.

## 5 EXPERIMENTAL RESULTS

Tests were performed to assess the system regarding its performance, temperamental characteristics and emotional behaviors analysis.

To perform our tests, we evaluated the agent's performance on reaching the goal. We also evaluated the appraisal values modifications during the simulation time. We performed the evaluation of an entire team of nine agents, in order to compare their performance with other teams of agents. During these evaluations we tried to analyse the difference between distinct temperamental teams and compare them in general terms (PAD scale and emotion valence), as well as their performance on reaching the goal. We perform the evaluation on three different simulation scenarios:

- First scenario has few obstacles (walls) and a small arena. These conditions enable fast detection of the beacon to the agents, but force the interaction between agents across the simulation arena.
- Second scenario has a lot of obstacles (walls) and a larger arena. In this scenario the goal is very difficult to accomplish and the agents have enough space for team interaction (grouping or isolation).
- Third scenario has a lot of obstacles (walls) and a small arena. In this scenario the goal is very difficult to accomplish, and the agents have fewer space for team interaction (grouping or isolation).

To perform our tests we have defined the same simulation time for each team (180 seconds), but if some agent have accomplished the goal during simulation cycle, his simulation time is considered as a time he spend to accomplish the goal. For example, if we have some team with two agents and one of them has accomplished the goal in 200 seconds and other haven't accomplished the goal, the medium time of the team will be calculated as (1800+200)/2=1000 seconds and the best time will be considered as 200 seconds.

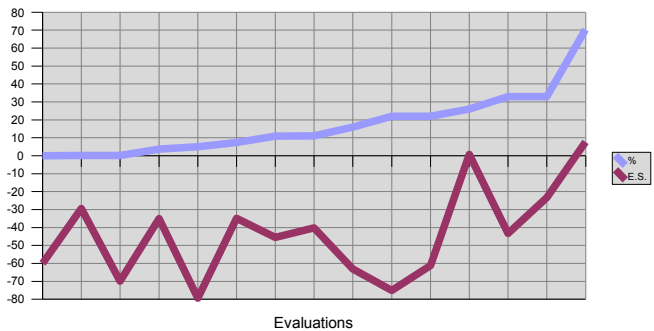

**Figure 7**: Team Performance vs Appraisal emotional state

We analyse our agent's performance on these three simulation scenarios and try to discover the advantages and disadvantages of using temperamental agents with this kind of simulation. We perform 10 interactions for each team/simulation scenario. The PAD values are presented in [-10, 10] interval instead of [-1, 1].

We perform the tests with Choleric, Sanguine, Melancholic and Phlegmatic homogeneous teams and with heterogeneous multi-temperamental team of agents.

Figure 7 shows the dependencies between Team Performance and Emotional State of the agents. In our architecture the performance of the agents doesn't depend on appraisal mechanism which only controls the psychical layer of the agent and only influences his PAD values and the emotional state. The agents performance only depends on temperamental (physiological) configuration of the agent (motors, sensors, anxiety, etc..) and his decision layer based on extrovert/introvert characteristics. So, we can see that the temperamental decision mechanism clearly influence the emotional state of the agent during the simulation.

In order to explain the nature of this influence we think that the implementation of the Pavlov's temperamental theory results in agents executing different grouping activities to satisfy their temperamental needs, so, this could influence the PAD values and consequently the Emotional State of the agent. Also we can analyse the influence of the system goals on the Emotional State of the agent (from the Appraisal Bank), and as we describe in Section 4, the decision temperamental mechanism works in order to accomplish some of these goals (avoid the walls or reach the beacon, for instance). So, even having no direct dependence between the implementation of these two layers, there are similar goals which are defined and this could explain the dependence between the agent performance and his Emotional State.

## 6 CONCLUSIONS AND FUTURE WORK

A main goal of this project was to develop and test a new model of a computational emotional mind using two different temperamental theories. For tackling this domain, we first had to study the basics of psychology and different approaches to evaluate emotional life of personality from temperamental perspective. As a rule, the theories of emotions can say too little about the role of emotions in the development of personality and about their influence on thought and action. The majority of the researchers of emotions are connected only with one of the components of the emotional process. Although some theories develop the separate aspects of the interrelations of emotion and reason, actions and personality, much still must be done both on the theoretical and on the empirical levels.

In this work we have tried to implement emotional agents using two different approaches: appraisal theory with the PAD model and the central nervous system theory. We approach the concept of emotion from the physiological and psychical perspective, defining the personality of the agent and analysing the different components of agent behaviors. We have simulated a kind of homogeneous and non-homogeneous society with different personalities and analysed their group and individual performances.

We can conclude that our approach produce very good results showing the dependence between two different layers (physiological and psychical) which where implemented independently. So, as it already has been proved theoretically from psychological perspective, which define that our emotional process are dependent on our temperamental type, we could state that our architecture is consistent and show the same dependence between two layers. This let us a large room for future improvement and research on this area.

The modular approach to emotion programming is a very promising way to integrate emotions into multi-agent systems with different goals and configurations. Temperament helps to support agent's decision making and with proper use can improve the agent's performance and the global teamwork. Also, our system helps us analyse the configurations we could choose to implement the personality in the system with different and particular characteristics, helping us to select the variables and functions of personality with better fitness to the specific system.

We are planning to implement in this project different search algorithms for evaluate the impact of emotions and temperament on search strategies. Other development is the introduction of visual emotional feedback using the face expressions such as proposed by the Russel [19]. Also we are aiming at the introduction of additional objects in the simulation environment with different degree of thread/satisfaction.